%% file: main.tex
\documentclass[10pt,twocolumn,letterpaper]{article}

\usepackage[pagenumbers]{cvpr} % To force page numbers, e.g. for an arXiv version

\usepackage{graphicx}
\usepackage{amsmath}
\usepackage{amssymb}
\usepackage{dsfont}
\usepackage{booktabs}
\usepackage{bbding}
\usepackage{makecell}

\usepackage{multicol}
\usepackage{multirow}
\usepackage[ruled,vlined]{algorithm2e}
\usepackage[dvipsnames]{xcolor}

\usepackage[pagebackref,breaklinks,colorlinks]{hyperref}

\usepackage{xspace}
\newcommand{\NMSE}[0]{NMSE\xspace}
\newcommand{\name}[0]{CLKD\xspace}
\newcommand{\ModelName}[0]{CLKD\xspace}

\usepackage{xcolor}

\usepackage[capitalize]{cleveref}
\crefname{section}{Sec.}{Secs.}
\Crefname{section}{Section}{Sections}
\Crefname{table}{Table}{Tables}
\crefname{table}{Tab.}{Tabs.}

\def\cvprPaperID{10278} % *** Enter the CVPR Paper ID here
\def\confName{CVPR}
\def\confYear{2023}

\begin{document}
\definecolor{commentcolor}{RGB}{110,154,155}   % define comment color
\newcommand{\PyComment}[1]{\ttfamily\footnotesize\textcolor{commentcolor}{\# #1}}  % add a "#" before the input text "#1"
\newcommand{\PyCode}[1]{\ttfamily\footnotesize\textcolor{black}{#1}} % \ttfamily is the code font

%\title{Knowledge Distillation is Born to be Asymmetrical Learning }
 
\title{Class-aware Information for Logit-based Knowledge Distillation}

\author{Shuoxi Zhang, Hanpeng Liu\\
School of Computer Science and Technology\\
Wuhan, China\\
{\tt\small \{zhangshuoxi,hanpengliu\}@hust.edu.cn}
\and 
John E. Hopcroft\\
Department of Computer Science, Cornell University\\
{\tt\small jeh@cs.cornell.edu}
\and
Kun He\\
\thanks{Corresponding author.}
School of Computer Science and Technology\\
Wuhan, China\\
{\tt\small brooklet60@hust.edu.cn}
}
\maketitle

%%%%%%%%% ABSTRACT
\begin{abstract}
Knowledge distillation aims to transfer knowledge to the student model by utilizing the predictions/features of the teacher model, and feature-based distillation has recently shown its superiority over logit-based distillation. However, due to the cumbersome computation and storage of extra feature transformation, the training overhead of feature-based methods is much higher than that of logit-based distillation. In this work, we revisit the logit-based knowledge distillation, and observe that the existing logit-based distillation methods treat the prediction logits only in the instance level, while many other useful semantic information is overlooked. To address this issue, we propose a Class-aware Logit Knowledge Distillation (CLKD) method, that extents the logit distillation in both instance-level and class-level. 
CLKD enables the student model mimic higher semantic information from the teacher model, hence improving the distillation performance. We further introduce a novel loss called Class Correlation Loss to force the student learn the inherent class-level correlation of the teacher.  
Empirical comparisons demonstrate the superiority of the proposed method over several prevailing logit-based methods and feature-based methods, in which CLKD achieves compelling results on various visual classification tasks and outperforms the state-of-the-art baselines.  
\end{abstract}

%%%%%%%%% BODY TEXT
\section{Introduction}\label{intro}
\input{body/intro}

\section{Related Work}\label{RW}
\input{body/RW}

\section{Methodology}\label{method}
\input{body/method}

\section{Experiments}\label{Experiments}
\input{body/Experiments}

\section{Further Discussion}\label{Discussion}
\input{body/discussion}
%Please follow the steps outlined below when submitting your manuscript to the IEEE Computer Society Press. This style guide now has several important modifications (for example, you are no longer warned against the use of sticky tape to attach your artwork to the paper), so all authors should read this new version.

\section{Conclusion}
In this work, we propose a novel logit-based distillation method to capture the inter-instance as well as  intra-instance knowledge by simple class representation. Our class representation module may exploit the latent relationship information across the instances. Our proposed method CLKD achieves state-of-the-art performance on various visual classification tasks. Given its success on classification tasks, we assume 
that our method would also work for more complicated visual tasks such as object detection and semantic segmentation. %However, due to a lack of context-structural knowledge transferring, our approach underperforms some of the state-of-the-art distillation methods designed for dense prediction. 
In our future work, we will study how to learn context-structural features for dense prediction and adopt our method on more complicated visual tasks.

%%%%%%%%% REFERENCES
{\small
\bibliographystyle{ieee_fullname}
\bibliography{main}
}
\clearpage
\appendix
\section*{Appendix}
\input{body/A}
\end{document}

%% file: body/intro.tex
%In the current wave of artificial intelligence in full swing, deep learning as an effective model tries to simulate the learning process of human beings. 
%In recent decades, deep learning revitalizes the field of computer vision, which has seen significant improvements in performance and generalization in various downstream tasks, such as image classification~\cite{he2016deep,hu2018squeeze}, object recognition~\cite{he2017mask,girshick2015fast}, image segmentation~\cite{zhao2017pyramid}, \etc However, with the improvements of performance, the models become larger and larger and are hardly deployed on mobile devices. To apply deep learning models on these lightweight devices, knowledge distillation~\cite{hinton2015distilling,romero2014fitnets,kim2018paraphrasing,zagoruyko2016paying,tian2019contrastive,chen2021cross,yue2020mgd} is used as a mechanism to compress large pre-trained models. % is gradually entering the research landscape.  

The great success of deep learning generally relies on the over-parameterized network to extract the representative features. Thus, deep learning models are usually too cumbersome to be deployed in mobile devices. It gives rise to a ground-breaking research field on \textbf{K}nowledge \textbf{D}istillation~\cite{hinton2015distilling,romero2014fitnets,kim2018paraphrasing,zagoruyko2016paying,tian2019contrastive,chen2021cross%,yue2020mgd
}~(KD), that aims to transfer knowledge from a complicated, pre-trained teacher network to a lightweight student model without sacrificing the performance. 
%transfer the \textit{dark knowledge}, referred to the hidden knowledge encapsulated in the secondary probabilities in the predictions, from a high-capacity teacher network to a compact student network. 
Approaches of KD fall into two categories, namely logit-based distillation and feature-based distillation.  

The key idea behind the vanilla KD~\cite{hinton2015distilling}, which belongs to logit-based distillations, is that the soft prediction probabilities contain more information than the ground-truth labels alone thus could help maintain the performance of the student model after distillation. 
Feature-based distillations~\cite{chen2021cross,zagoruyko2016paying,kim2018paraphrasing,yue2020mgd} extract intermediate features of the teacher model to be distilled to the student, and surpass logit-based distillations on various tasks owing to the flexibility of feature representation, and has become the mainstream KD in recent years.

%For more information about logit-based and feature-based distillation, the readers are referred to \cref{RW}. 

% However, a burning challenge in feature-based distillation is the inevitability of the representation gap between the teacher and the student since they have different capacities. Besides, it is challenging to choose which kinds of representation and which kinds of teacher networks should be transferred. Therefore, feature-based distillation is always along with tedious manual work. 
\begin{figure*} [t!]
	\centering
	\includegraphics[width=0.95\textwidth]{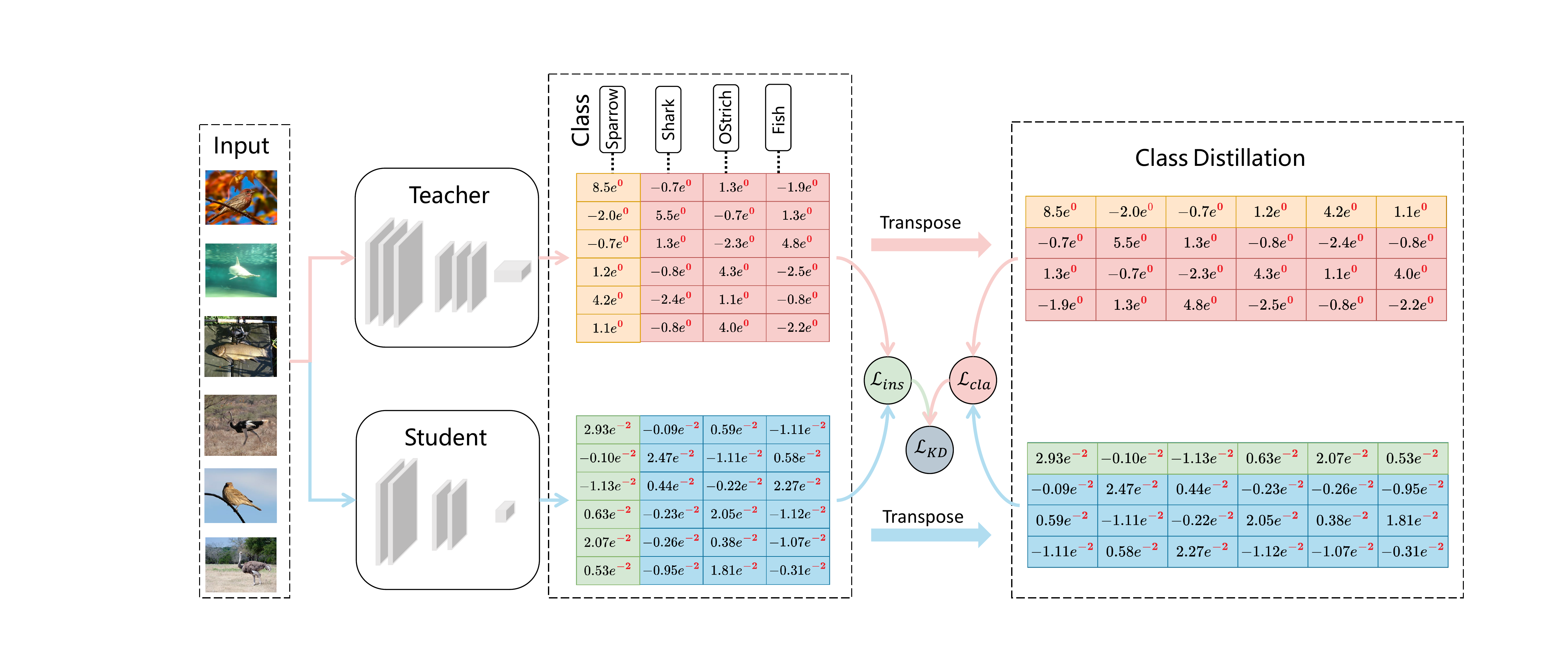}
	\caption{%The framework of the proposed \name method. 
    Illustration of the proposed \name framework. 
    The rows of each logit matrix are the prediction probabilities of the instances. The columns can be viewed as class representations embodied from instances. As a result,  instance-level and class-level contrastive learning can be conducted in 
         %the row space and column space of the feature matrix, respectively. 
         the feature matrix's row and column space, respectively.}
	\label{fig:framework} 
\end{figure*}

However, logit-based distillation should achieve comparable performance since logits are at a higher semantic level than intermediate features. Thus, the potential of logit-based distillation may still need to be fully explored. Recently, there are also some researches studying the weakness of logit-based distillation and proposing improvements to extract information from logits~\cite{zhao2022decoupled,mirzadeh2020assistant}. However, we notice that they only consider the instance-level information, and we argue that the instance-level information alone is not enough to transfer some vital semantic information, such as the \textit{inter-instance knowledge}. 

%We borrow the idea from \cite{li2021contrastive} to leverage the inter-instance knowledge.
As shown in \cref{fig:framework}, the image batch is fed into the teacher and student networks to output the prediction matrices. Rows of the matrices corresponds to the class probabilities that the network attempts to determine to which class the instance belongs. Columns of the matrices can be viewed as the \textit{class representation}, and each entry of a single column could represent the similarity between the class and the input instance, epitomizing the inter-instance knowledge. In this way, the \textit{inter-instance knowledge} and the \textit{intra-instance knowledge} could be obtained accordingly.  

Based on the above observation, we introduce a simple yet effective logit distillation method, termed class-
aware Logit Knowledge Distillation (\name), that blends inter-instance and intra-instance knowledge and allows the class correlation information to be embraced by class representation. As shown in \cref{fig:framework}, we first use the logit matrix transposition to extract inter-instance knowledge, which is complementary to intra-instance information, then we employ a class correlation loss to let the student mimic the teacher's class correlation. %, and the details will be discussed in \cref{md:dcor}. 
Besides, from \cref{fig:framework}, we observe that the amplitude gap of predictions between teacher and student is huge. Hence, we design a normalized metric to calculate the discrepancy between teacher and student to boost the distillation. Armed with these mechanisms, we could fully exploit the inter-instance and intra-instance information, making the distillation more effective.

To demonstrate the effectiveness of our approach, we conduct extensive experiments on standard benchmark datasets and  witness that \name consistently outperforms all the state-of-the-art methods we compared with, including prevalent feature distillation methods. Besides, since intermediate features also include inter-feature and intra-feature knowledge, we can adopt our approach to feature-based distillation to make further improvement. 
%And for each point considered, we make corresponding ablation experiments to show its effectiveness.

%In this paper, we propose the following: firstly, we combine the intra-instance and inter-instance knowledge by reusing the logit matrix. Secondly, we introduce class decorrelation to decrease redundancy somewhat. Finally, we use normalization to make the output of the teacher model and the student model comparable.

Our main contributions can be summarized as follows:
\begin{itemize}
\setlength{\itemsep}{0pt}
\setlength{\parsep}{0pt}
\setlength{\parskip}{0pt}
    \item We propose reusing the logit matrix so as to combine the intra-instance and inter-instance knowledge. The inter-instance knowledge obtained by transposing the logit matrix can reduce the gap between the teacher and the student models.
    \item We introduce a class correlation loss to mimic the inherent class relationship of the teacher. And we use normalization to make the output of the teacher and student models more comparable.
    \item Our approach consistently achieves superiority over the state-of-the-art baselines on extensive experiments, considering different network architectures and various tasks (classification, segmentation, feature transferring).
\end{itemize}

% 先介绍下cv在知识蒸馏方面的发展，然后再介绍下logit-base 和 feature-base的不同，提出我们的movtation。 考虑类间具有相似，不同的类应该相应的保持一定的距离。最后考虑到不同的类的 引入会增加计算量，设计了新的loss计算方法，引言的参考文献引用有点少。

%% file: body/RW.tex
The concept of KD was first proposed by Hinton \etal~\cite{hinton2015distilling}. KD directs the student training by leveraging the dark knowledge of teacher model, and enhances the performance of student model successfully. Dark knowledge, which can provide additional information to supervise the training process compared to simply utilizing ground-truth labels, is obtained from teacher networks in features or soft logits. Therefore, the studies on KD can be divided into two categories, \ie, logit-based distillation and feature-based distillation. 

\subsection{Logit-based Distillation}\label{logits distillation}
Hinton \etal introduce a temperature scaling method on prediction logits to distill the teacher's dark knowledge in the category similarity probabilities. In the tenets of logit-based distillation, the similarity probabilities from the negative logits may provide extra supervision and better regularization~\cite{yun2020regularizing}. Thus, logit-based distillation improves the performance and generalization of student model; however, there is still a big gap between teacher and student. Several subsequent works attempt to close the gap. Hedge \etal~\cite{hegde2020variational} improve the logit distillation by inducing variational inference. Mirzadeh \etal~\cite{mirzadeh2020assistant} introduce an extra small-size assistant network to close the gap between teacher and student. Zhao \etal~\cite{zhao2022decoupled} decouple the classical KD loss so as to enable the distillation loss to play their roles more %efficiently
effectively and flexibly. Kim \etal~\cite{kim2021comparing} propose a logit matching strategy for distillation and show its superiority over conventional Kullback-Leibler divergence~\cite{joyce2011kullback} on temperature-scaled logits. Chen \etal~\cite{chen2022reuse} reuse the teacher's classifier to mitigate the representation gap between teacher and student, and achieve the best performance so far; however, introducing the teacher's classifier aggravate the deployment costs, and it may violate the tenets of knowledge distillation. Besides, the problem of performance gap is still far from being solved. 
%these approaches design different mechanisms to improve the logit distillation, the problem of performance gap is far from being solved. 

\subsection{Feature-based Distillation}
\label{feature distillation}
To alleviate the performance gap between teacher and student in logit distillation, considerable researches on knowledge distillation concentrate on leveraging the feature information embedded in intermediate layers. Rather than distilling the prediction logits, feature-based distillation aims to force the student to mimic the feature representations of the teacher. Feature distillation is first introduced in FitNet~\cite{romero2014fitnets}, which utilizes the {`hint'} to force the intermediate features of student to mimic the corresponding part of teacher. Inspired by FitNet, a variety of methods have been proposed to match the features. To be specific, attention map~\cite{zagoruyko2016paying}, neural selectivity~\cite{huang2017like} and paraphrased information~\cite{kim2018paraphrasing} of the teacher network are proposed to express the knowledge. 
%introduce a ``attention map'' from the original feature maps to transfer the knowledge. \cite{huang2017like} transfer the attention map via neural selectivity. \cite{kim2018paraphrasing} propose the alleged \textit{factors of the network} as a more transferable form of intermediate representations. 
Feature-based distillation induces richer information and provides higher flexibility for the knowledge transfer. However, due to the difference in feature size between teacher model and student model, feature-based distillation requests adding extra layer transformation to align the different sizes. Therefore, the additional computation %costs and memorial storage 
and memory overhead would affect the %popularity and use 
wide application 
of feature-based distillation.

%To avoid extra costs in feature distillation, we need to revisit the logit-based distillation. 
Therefore, we revisit the logit-based distillation in this work and wish to take fully utilization of its potential on KD. 
Intuitively, logits are at a higher semantic level than intermediate features, and logit-based distillation should achieve comparable performance. We focus on what limits logit distillation and propose a novel method that %may revive logit distillation. 
gain comparable or even higher performance than the state-of-the-art baselines. 

To some extent, our key idea is related to the previous study on \textbf{C}ontrastive \textbf{C}lustering (CC)~\cite{li2021contrastive}. CC introduces the cluster representation in instance-wise contrastive learning to avoid mode collapse, which is the primary concern in contrastive learning~\cite{chen2020simple,he2020momentum}. Different from CC, our method purposes to reduce the performance gap in the KD paradigm, rather than alleviating the mode collapse which is not the main cause of performance degrade in KD. In addition, compared to adding clustering projections in CC, our model makes no necessity of adding extra branches in the framework of logit distillation.

%Besides logit distillation, in recent years, various feature-based distillation approaches have been proposed. These works mainly attempt to directly match the feature representation of the teacher and the student. Though feature-based knowledge transfer provides preferable information for the learning of the student model. However, due to the significant differences between sizes of the hints layer from the teacher model and the guided layers from the student, adding extra transform layers become inevitable. In consequence, training cost of feature distillation methods are unbearable.

%% file: body/method.tex
In this section, we first briefly review %the preliminary of 
the vanilla \textbf{K}nowledge \textbf{D}istillation (KD). Second, to grasp the teacher knowledge more %efficiently
effectively, we %utilize the %matrix transposition on the output matrix 
transpose the output matrix to capture the class-instance dependency. To mitigate the amplitude gap of the outputs between teacher and student, we utilize a novel metric for distillation, termed the Normalized Mean-Squared Error (\NMSE). Then, inspired by the idea behind Correlation Congruence~\cite{peng2019correlation}, we further propose a strategy to grasp the class correlation information learned by the teacher. 
%and we will show its superiority over KL divergence and Mean-Squared Error (MSE)~\cite{kim2021comparing}. 
In the end, we show that the proposed method can also be applied to feature-based distillation.

\subsection{Vanilla Knowledge Distillation}
Here we briefly recap the basic %tenets
idea of vanilla knowledge distillation~\cite{hinton2015distilling} and provide necessary notations for knowledge distillation. 
%Given a dataset $\mathcal{X}=\{\boldsymbol{x_1},\boldsymbol{x_2},\cdots,\boldsymbol{x_N}\}$ from $K$ categories with corresponding labels $\mathcal{Y}=\{\boldsymbol{y_1},\boldsymbol{y_2},\cdots,\boldsymbol{y_N}\}$, where $N$ denotes
In the vanilla KD, information from the teacher is epitomized and transferred in the form of softened logit probabilities. The total loss can be expressed as:
\begin{equation}\label{eq:kd loss}
\mathcal{L} = (1-\alpha)\mathcal{L}_{CE}(\sigma(\boldsymbol{z}_S),\boldsymbol{y}) + \alpha\mathcal{L}_{KD}(\boldsymbol{z}_S, \boldsymbol{z}_T),
%\mathcal{L} = (1-\alpha)\mathcal{L}_{CE}(\sigma(\boldsymbol{z}_S),\boldsymbol{y}) + \alpha\mathcal{L}_{KL}
%\alpha{\tau}^2\mathcal{L}_{KL}\left(\sigma(\frac{\boldsymbol{z}_S}{\tau}), \sigma(\frac{\boldsymbol{z}_T}{\tau})\right)
\end{equation}
where $\mathcal{L}_{CE}(\cdot,\cdot)$ denotes the cross-entropy classification loss, $\sigma(\cdot)$ is the softmax function and $\boldsymbol{y}$ the one-hot vector indicating the corresponding label. The distillation loss $\mathcal{L}_{KD}(\cdot, \cdot)$ is calculated by using $\boldsymbol{z}_S$, $\boldsymbol{z}_T$, \ie, the predictions of student and teacher, respectively. The hyper-parameter $\alpha \in [0,1]$ weights the knowledge distillation loss and cross-entropy loss. In the vanilla KD, Hinton \etal utilize the temperature-scaled Kullbuck-Leibler divergence to enforce the student mimic the softened categorical probability distribution of the teacher:
\begin{equation}\label{eq:kl loss}
%\mathcal{L}_{KD}={\tau}^2\mathcal{L}_{KL}\left(\sigma(\frac{\boldsymbol{z}_S}{\tau}), \sigma(\frac{\boldsymbol{z}_T}{\tau})\right),
\mathcal{L}_{KD}={\tau}^2\mathcal{L}_{KL}\big( \sigma(\boldsymbol{z}_S/\tau), \sigma(\boldsymbol{z}_T/\tau) \big),
\end{equation}
where the temperature $\tau$ is used for logit softening. When $\tau$ equals $1$, %$\sigma(\frac{\boldsymbol{z}}{\tau})$ 
$\sigma(\boldsymbol{z}/\tau)$  becomes the standard softmax function. When $\tau$ increases, the probability distribution becomes softer, providing extra information as to which classes are more similar to the ground-truth class in the teacher's mind. 
%In practice, for a mini-batch of instances in size $B$, 
%the outputs of the teacher and student networks satisfy that 
%$\boldsymbol{z}_T, \boldsymbol{z}_S \in \mathbb{R}^{B\times{C}}$, where $C$ is the number of categories. 

\subsection{Categorical Level Knowledge Distillation}
The vanilla KD assumes that the teacher's prediction on a single image would carry the instance-level class similarity knowledge that can be transferred by distillation. In this work, we notice that the information in the teacher's prediction has not been fully exploited because the vanilla KD treats the prediction outputs only in the instance level. In this subsection, we %decouple the output matrix into two parts, 
aim to learn from the output matrix from two perspectives.  
One is from the instance-wise probability distribution, the other is from categorical representations which may transfer knowledge at a higher semantic level. 
 
\noindent \textbf{Motivation.}  
%The model outputs, $\boldsymbol{z}_T$ and $\boldsymbol{z}_S$, are matrices with $B$ rows (batch size) and $C$ columns (number of classes). From the algebraic view, The logit matrix consists of row vectors, which are the instance-level categorical probability prediction. The vanilla KD captures categorical probability information instance-wisely by simulating the student's output matrix to the teacher's row-wisely. However, only considering the instance-level distillation neglects the information in the logit matrix's column, which carries the class-level information across the instances. Thus, the vanilla KD and other prevailing logit distillation methods only construct the relationships \textit{within a single image}, regardless of the semantic dependencies across different images.  
For a mini-batch of instances in size $B$, the logit matrices $\boldsymbol{z}_T, \boldsymbol{z}_S \in \mathbb{R}^{B\times{C}}$, where $C$ is the number of categories. The row vectors are instance-level categorical probability prediction. The vanilla KD captures the categorical probability information by simulating the teacher's logit matrix row-wisely.  However, the column vectors of the logit matrix actually contain rich class-level information across the instances.  While the existing prevailing logit distillation methods only construct the relationships \textit{within a single image}, but overlook semantic dependencies \textit{across different images}.  

To this end, we propose a novel KD method that exploits the logit matrix beyond the instance-wise level. Specifically, we retrieve both the instance-wise information and the class-wise information %by decoupling 
from the logit matrices of teacher and student, thus transferring the teacher's knowledge more comprehensively.
%a more holistic knowledge. %Details of our approach is presented in the following.

\subsubsection{Class Distillation}\label{md:cluster}
 Given a mini-batch of data $\{\boldsymbol{x}_n\}_{n=1}^B$, we pick the outputs %($B\times{C}$) 
 before the softmax layer as our logit matrix. 
 %To distill the cluster information, we view 
 We then regard each of its column vector as the representation of the corresponding class. 
 Each entry of the column vector represents the similarity between this class and the corresponding instance, which is the class-instance similarity knowledge. As shown in \cref{fig:framework}, the critical component of our method is the ``\textit{class distillation}''.
 %, \ie,  we maneuver the class-wise similarity information by using simple transposition. 
 Through matrix transposition, the inter-instance information could be transferred by forcing the student to produce a similar class representation to the teacher's. The loss function is defined as: %can be expressed by:
\begin{equation}\label{eq:total loss}
\begin{aligned}
    &\mathcal{L}_{KD} = \mathcal{L}_{ins} + \beta\mathcal{L}_{cla}, \\ 
    &\mathcal{L}_{ins} = \mathcal{L}_{\NMSE}(Z_s, Z_t), \\ % \textrm{Loss}(Z_s, Z_t) \\
    &\mathcal{L}_{cla} = \mathcal{L}_{\NMSE}(\textrm{norm}^T(Z_s), \textrm{norm}^T(Z_t)), \\
    &\mathcal{L}_{\NMSE}(p,z) = \left\lVert{\frac{p}{\lVert{p}\rVert_2} - \frac{z}{\lVert{z}\rVert_2}}\right\rVert_2^2,
    %\textrm{Loss}(\textrm{Normlize}(Z_s)^H, \textrm{Normlize}(Z_t)^H),
\end{aligned}
\end{equation}
where $\mathcal{L}_{ins}$ is the conventional \textbf{ins}tance-wise distillation loss, and $\mathcal{L}_{cla}$ is our proposed \textbf{cla}ss-wise distillation loss. %The total loss includes the $\mathcal{L}_{ins}$ loss and the $\mathcal{L}_{ins}$ loss traded off by coefficient $\beta$. 
Coefficient $\beta$ is used to trade off the two terms. $p$ and $z$ denote two different encodings or distributions, which in our case are the logits of teacher and student, respectively.
The normalization ($\ell_2$ norm) before matrix transposition is essential for alleviating the impairment to the class representation, due to the amplitude gap between different rows. Kim \etal~\cite{kim2021comparing} propose the logit matching strategy (MSE between logits) for distillation and show the priority of logit matching over conventional KL divergence. However, due to the inevitable gap in model size and parameter number between teacher and student, the norm of outputs can hardly share a similar amplitude, restricting the simulation of simple logit matching. Therefore, we normalize the logits before calculating the MSE loss and propose a novel metric, called \NMSE, to calculate the difference between teacher and student during the distillation.  In \cref{dis-abl:metrics}, we will test the performance when using other metrics to measure the distribution discrepancy between teacher and student, such as MSE, KL divergence, \etc. 

\begin{figure} [t!]
	\centering
	\includegraphics[width=0.5\textwidth]{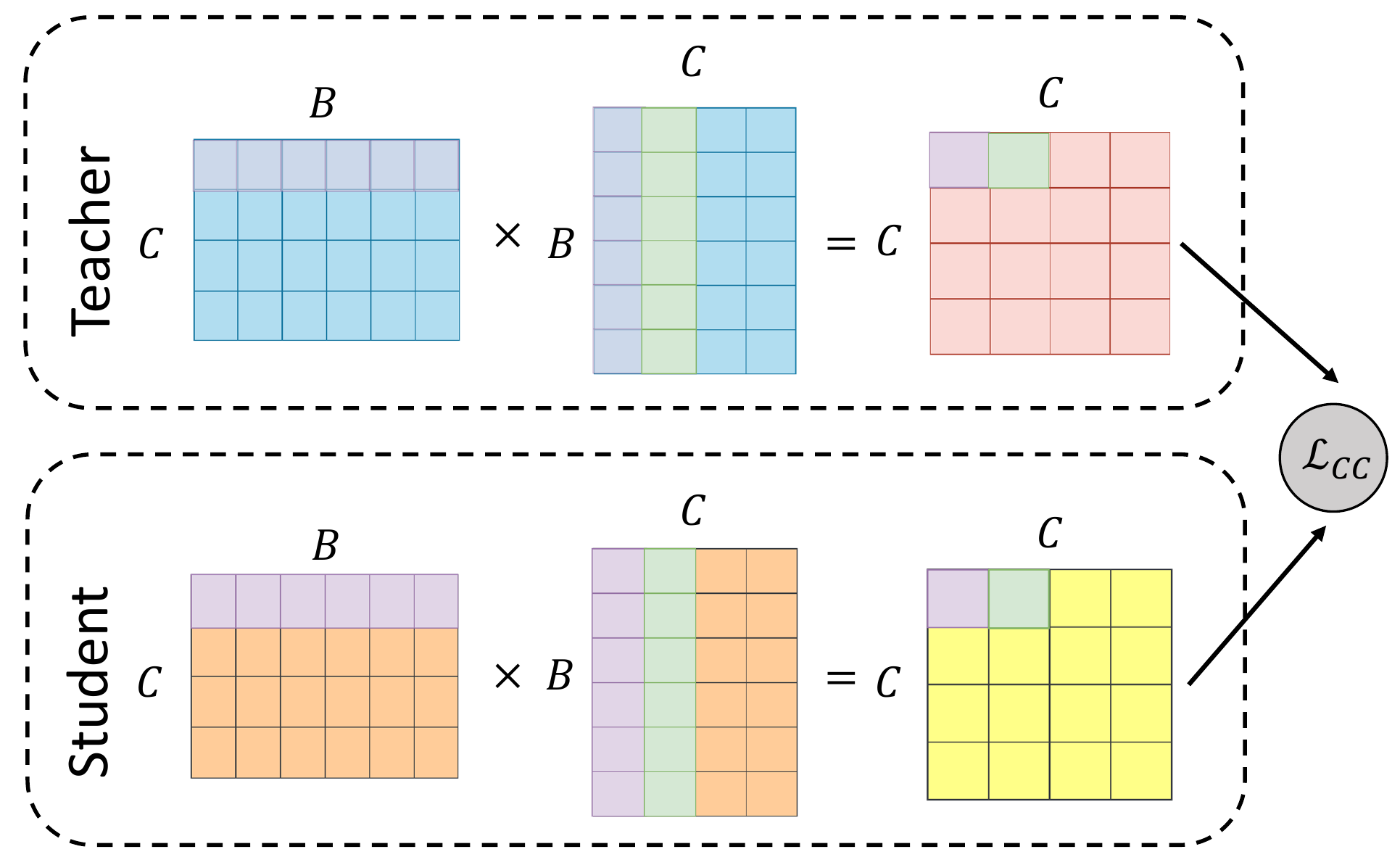}
	\caption{We use class correlation distillation to guide the student to learn the class correlation information of the teacher. Given the $(B\times{C})$ logit matrices, we develop $C\times{C}$ class correlation matrices, and compute the difference of the correlation matrices between the teacher and the student.}
	\label{fig:correlation} 
\end{figure}

\subsubsection{Class Correlation Loss}\label{md:dcor}
As we encode the class representation in logit transposition, our class representation may manifest the correlation between different classes. The question is: how can we learn the class correlation of the student? Inspired by the idea of Correlation Congruence~\cite{peng2019correlation} to capture the instance correlation, we attempt to force the student's class correlation simulate the teacher's. Hence, we calculate the class correlation matrix of teacher and student, respectively. The class correlation matrix has the following form:
\begin{equation} \label{eq:correlation matrix}
\begin{aligned}
    \mathcal{B}(Z) &= \frac{1}{C - 1} \sum_{j=1}^{C} (Z_{\cdot{j}} - \bar{Z})^T(Z_{\cdot{j}} - \bar{Z}), \\ 
    %&\textrm{where} \ \ \ \bar{Z}_{\cdot} = \frac{1}{C} \sum_{j=1}^{C} Z_{ij}.
\end{aligned}
\end{equation}
where $\bar{Z}\in \mathbb{R}^C$ denotes the mean vector across columns and $Z_{\cdot{j}}$ is the $j$-th column vector. We force the computed class correlation of the student to resemble the teacher's. Thus the {\bf C}lass {\bf C}orrelation (CC) loss is as shown in \cref{eq:cov_loss}, calculated by the element-wise sum of the discrepancy of the class correlation matrices of teacher and student. %The decorrelation loss can be written as:
\begin{equation} \label{eq:cov_loss}
    \mathcal{L}_{CC}(S, T) = \frac{1}{C^2}\lVert{\mathcal{B}(Z_S)-
    \mathcal{B}(Z_T)}\rVert_2^2,
\end{equation}
where $\mathcal{B}(Z_S)$ and $\mathcal{B}(Z_T)$ are the correlation matrix of student and teacher, respectively. The class correlation loss encourages the student to capture the class correlation information of the teacher. To sum up, the overall loss function is weighted by the teacher-student KD loss, as well as the classification loss and the correlation loss:
\begin{equation} \label{eq:loss}
    \mathcal{L} = \ \lambda \mathcal{L}_{CE} + \mu\mathcal{L}_{KD} + \nu\mathcal{L}_{CC}, \\
\end{equation}
where $\lambda, \mu, \nu \in [0,1]$, $\lambda+\mu+\nu=1$ are hyper-parameters controlling the importance of each term. We illustrate the architecture and loss function of our approach in \cref{fig:framework}.

\subsection{Feature Matrix Transposition}
Feature distillation exploits the knowledge in intermediate layers, thus adding the flexibility of the teacher's knowledge. 
%We may wonder that can we merge our approach into feature-base distillation. 
In this subsection, we aim to adapt our class-aware logit-based distillation to the feature distillation to grasp the feature information more properly. 
Generally speaking, the conventional feature-based distillation loss can be formulated as: 
\begin{equation}
    \mathcal{L}_{FeaKD}\left(F_s(x), F_t(x)\right) = \mathcal{L}_{Fea}\left(\Phi_s(F_s(x)),\Phi_t(F_t(x))\right),
    \label{feature distillation}
\end{equation}
where $F_s(x)$ and $F_t(x)$ are intermediate feature maps of student and teacher model, respectively. The transformation functions, $\Phi_s(F_s(x))$ and $\Phi_t(F_t(x))$, are usually applied for aligning feature maps of student and teacher with different shapes. And $\mathcal{L}_{Fea}$ is the distance function to match their feature maps. Similar to our logit-based distillation, feature-based distillation preserves information of the features, \ie, the instance-wise features. We attempt to utilize the cross-instance information on feature-based distillation. First, the aligned feature map matrices in $\mathbb{R}^{B\times{D}}$, are matrices in batch size $B$ and dimension $D$ for each instance. In our logit distillation, the encoded class representation may hint the relationship between instances, thus the cross-instance feature representation may also imply the similarity of instances. For example, the intermediate feature may have various dimensions, which may represent different representations of the image. 
Each of these features can be symbolized by different instances, 
%just like the matrix transposition we conduct in our logit distillation. 
 which is similar to 
 the class-wise distillation loss in logit distillation. 
We could exploit the intra-instance and the inter-instance feature information similarly. We apply our approach on several feature-based distillation such as FitNet~\cite{romero2014fitnets}, AT~\cite{kim2018paraphrasing}, CRD~\cite{tian2019contrastive} and SRRL~\cite{yang2021softmax}; and in \cref{dis:feature distillation}, we will illustrate the effectiveness of applying our method to these  feature-based distillations.

%% file: body/Experiments.tex
To show the effectiveness of our approach, we compare \name with several state-of-the-art approaches by evaluating on several image classification tasks. We also evaluate our approach with the teacher-free distillation paradigm in the Appendix.  
%We also compare with several attention-based visual models in the Appendix. 
%The ablation study and parameter study are implemented, with details in the Appendix. 
%We evaluate our proposed \name method on several image classification tasks through a small, size-limited neural network by utilizing its own distillation. We then analyze how our method works by quantitative evaluations. In the end, we provide ablation studies for further analysis.

\subsection{Experimental Setup}\label{basic setting}
%Firstly, the experimental details are as follows:
%\\
\noindent {\bf Datasets.} We evaluate the effectiveness of \name on the CIFAR-100 dataset~\cite{krizhevsky2009learning}, which consist of ($32\times32$) color scaled images containing objects in 100 classes. We also test on ImageNet dataset~\cite{russakovsky2015imagenet} to show the efficiency on large-scale classification, of which all the images are resized to ($224\times224$). We operate the standard data augmentation and normalization as conducted in \cite{he2016deep,zagoruyko2016wide,huang2017densely} on the above two datasets.

%We further evaluate our method on several fine-grained datasets, including Caltech-UCSD Bird (CUB200)~\cite{WahCUB_200_2011}, Stanford 40~\cite{yao2011human}, Stanford Dogs (Dogs)~\cite{KhoslaYaoJayadevaprakashFeiFei_FGVC2011}, and MIT Indoor Scene Recognition (MIT67)~\cite{MIT67}. 
We conduct a series of classification with typical convolutional neural network architectures and lightweight networks on the CIFAR-100 and ImageNet datasets. Our implementation for CIFAR-100 follows the practice in~\cite{chen2021cross}. A variety of teacher-student pairs based on popular visual network architectures are tested, including ResNet~\cite{he2016deep}, VGG~\cite{simonyan2014very}, WideResNet (WRN)~\cite{zagoruyko2016wide}, and lightweight networks such as MobileNet~\cite{howard2017mobilenets} and ShuffleNet~\cite{zhang2018shufflenet}.

\noindent {\bf Training Details for CIFAR-100.} 
For the training of CIFAR-100, we adopt SGD optimizier with 0.9 Nesterov momentum, the total training epoch is predetermined to 240, and we divide the learning rate by 10 at epochs 150, 180, and 210. For the training, we use the standard data augmentation technique, \ie, flipping and random cropping. The initial learning rate is set to 0.01 for lightweight architectures, %such as MobileNets and ShuffletNets, 
and 0.05 for the other series. We train with weight decay $5e^{-4}$ for regularization. 

\noindent {\bf Training Details for ImageNet.} For the training of ImageNet, we follow the practice suggested by the PyTorch official. We train the models for 120 epochs. The initial learning rate is set to 0.1, and divided by 10 for every 30 epochs. The batch size is 512, and the weight decay rate is $1e^{-4}$. Besides, we deploy the cosine schedule with $5$-epoch warm-up for training. All experiments on the ImageNet dataset are trained on 8 GPUs for 120 epochs if not specified. The optimal results are chosen to maximize the top-1 accuracy on the validation set. 
%we apply two typical convolutional neural network architectures for small datasets, \ie, ResNet18~\cite{he2016deep} with 64 filters and DenseNet-121~\cite{huang2017densely} with a DenseNet-BC structure~\cite{huang2017densely}. We set both architectures  the first convolutional layer of the network with kernel size $3 \times 3$, strides 1 and padding 1 for image size $32 \times 32$.

%\noindent {\bf Hyperparameters.} All networks for small-scale datasets are trained from scratch and optimized by stochastic gradient descent (SGD) with momentum set to 0.9, weight decay $5e^{-4}$, and the initial learning rate 0.1. The learning rate is divided by 10 after 100 and 150 epochs respectively for each dataset, and the total epochs are 200. We set the batch size to 128 for conventional setting. We use the standard data augmentation technique, \ie flipping and random cropping. In our method, the temperature $\tau$ for KL divergence between intermediate layers and the penultimate feature is chosen from \{1,2,3,4\}, and the temperature $\tau_\alpha$ for softening attention alignment is selected from \{1,2,3,4\}. 

%\noindent {\bf Settings for ImageNet Dataset.} To demonstrate that our \name works well on large-scale classification tasks, we study the family of several visual networks, including ResNets, ResNeSts, ResNeXts on ImageNet. Our choice of temperature $\tau, \tau_\alpha$ is the same as our configuration on small datasets. The batch size of ImageNet training is set to 512. 

\noindent {\bf Baselines.} We compare our approach with two kinds of prevalent and advanced knowledge distillation baselines, \ie, logit-based distillation and feature-based distillation: 
\begin{itemize}
\setlength{\itemsep}{0pt}
\setlength{\parsep}{0pt}
\setlength{\parskip}{0pt}
	\item \textbf{Logit-based distillation} includes the vanilla KD~\cite{hinton2015distilling}, DTD-KA~\cite{wen2021preparing}, VBD~\cite{hegde2020variational} and 
	DKD~\cite{zhao2022decoupled}.
	\item \textbf{Feature-based distillation} includes FitNet~\cite{romero2014fitnets}, AT~\cite{zagoruyko2016paying}, SP~\cite{tung2019similarity}, VID~\cite{ahn2019variational}, MGD~\cite{yue2020mgd}, SRRL~\cite{yang2021softmax}, CRD~\cite{tian2019contrastive} and SemCKD~\cite{chen2021cross}. 
    %The detailed discussion is presented in \cref{exp:main results}. 
\end{itemize}

\begin{table*}[htbp]
	\centering
	\resizebox{0.95\textwidth}{!}{
		\begin{tabular}{l|l|cccccc}
			\toprule
			\multirow{2}{*}{Type} & \multirow{2}{*}{Student} & ResNet-8$\times$4 & VGG-8 & ResNet20 & WRN-40-1 & WRN-16-2 & ResNet32 \\
			& & 72.51 $\pm$ 0.29 & 70.46 $\pm$ 0.29 & 69.06 $\pm$ 0.22 & 71.98 $\pm$ 0.17 & 73.43 $\pm$ 0.22 & 71.14 $\pm$ 0.25  \\
			\midrule
			\multirow{2}{*}{---} &\multirow{2}{*}{Teacher} & ResNet-32$\times$4 & VGG-13 & ResNet56 & WRN-40-2 & WRN-40-2 & ResNet110  \\
			& & 79.42 & 74.64 & 73.44 & 76.31 & 76.31 & 74.31 \\
			\midrule
			\multirow{4}{*}{Logits} & KD \cite{hinton2015distilling} & 74.12 $\pm$ 0.15 & 72.66 $\pm$ 0.13 & 70.66 $\pm$ 0.22 & 73.42 $\pm$ 0.22 & 74.92 $\pm$ 0.20 & 73.02 $\pm$ 0.16 \\
			& DTD-KA \cite{wen2021preparing} & 73.78 $\pm$ 0.22 & 72.98 $\pm$ 0.14 & 70.99 $\pm$ 0.24 & 73.49 $\pm$ 0.16 & 74.73 $\pm$ 0.20 & 72.88 $\pm$ 0.13 \\
			& VBD \cite{hegde2020variational} & 74.31 $\pm$ 0.21 & 73.21 $\pm$ 0.17 & 71.13 $\pm$ 0.17 & 73.62 $\pm$ 0.22 & 75.10 $\pm$ 0.20 & 73.21 $\pm$ 0.22 \\
			& DKD \cite{zhao2022decoupled} & 76.32 $\pm$ 0.26 & 74.68 $\pm$ 0.23 & 71.79 $\pm$ 0.17 & 76.11 $\pm$ 0.17 & {\bf 76.55 $\pm$ 0.14} & 74.11 $\pm$ 0.17 \\
			\midrule
			\multirow{8}{*}{Features} & FitNet \cite{romero2014fitnets} & 73.89 $\pm$ 0.22 & 73.54 $\pm$ 0.12 & 71.52 $\pm$ 0.16 & 74.12 $\pm$ 0.20 & 75.75 $\pm$ 0.12 & 72.52 $\pm$ 0.07 \\
			& AT \cite{zagoruyko2016paying} & 74.57 $\pm$ 0.17 & 73.63 $\pm$ 0.12 & 71.76 $\pm$ 0.14 & 74.43 $\pm$ 0.11 & 75.28 $\pm$ 0.13 & 73.32 $\pm$ 0.11 \\
			& SP \cite{tung2019similarity} & 73.90 $\pm$ 0.17 & 73.44 $\pm$ 0.21 & 71.48 $\pm$ 0.11 & 73.17 $\pm$ 0.21 & 75.34 $\pm$ 0.21 & 73.63 $\pm$ 0.21 \\
			& VID \cite{ahn2019variational} & 74.49 $\pm$ 0.21 & 73.96 $\pm$ 0.17 & 71.71 $\pm$ 0.08 & 74.20 $\pm$ 0.18 & 74.79 $\pm$ 0.20 & 73.89 $\pm$ 0.19 \\
			& MGD \cite{yue2020mgd} & 74.41 $\pm$ 0.16 & 74.28 $\pm$ 0.21 & 71.68 $\pm$ 0.24 & 74.78 $\pm$ 0.22 & 75.98 $\pm$ 0.14 & 74.10 $\pm$ 0.14 \\
			& SRRL \cite{yang2021softmax} & 75.37 $\pm$ 0.12 & 74.68 $\pm$ 0.24 & 72.01 $\pm$ 0.27 & 74.98 $\pm$ 0.21 & 75.55 $\pm$ 0.14 & 74.21 $\pm$ 0.11 \\
			& CRD \cite{tian2019contrastive} & 75.59 $\pm$ 0.23 & 73.88 $\pm$ 0.18 & 71.68 $\pm$ 0.11 & 75.51 $\pm$ 0.22 & 76.01 $\pm$ 0.11 & 73.48 $\pm$ 0.16 \\
			& SemCKD \cite{chen2021cross} & {75.58 $\pm$ 0.22} & {74.42 $\pm$ 0.21} & {71.98 $\pm$ 0.17} & {74.78 $\pm$ 0.21} & {75.42 $\pm$ 0.15} & {74.12 $\pm$ 0.22} \\
			%\cmidrule{1-8}
			\midrule
			Logits & \name (Ours) & \textbf{77.68 $\pm$ 0.22} & \textbf{75.01 $\pm$ 0.12} & \textbf{72.71 $\pm$ 0.23} & \textbf{76.12 $\pm$ 0.21} & {76.32 $\pm$ 0.17} & \textbf{74.82 $\pm$ 0.28} \\
			
			\bottomrule
		\end{tabular}
	}
	\vspace{3pt}
	\caption{Top-1 test accuracy (\%) of various distillation approaches on CIFAR-100. The teacher and student pairs share similar architectures. Each experiment is repeated three times, and we report the mean and standard deviation of the top-1 accuracy. The best results appear in \textbf{bold}.
	}
	\label{Tbl:CIFAR-100-1}
\end{table*} 

\vspace{3pt}

\begin{table*}[t]
	\centering
	\resizebox{0.95\textwidth}{!}{
		\begin{tabular}{l|l|cccccc}
			\toprule
			\multirow{2}{*}{Type} & \multirow{2}{*}{Student} & ShuffleV1 & WRN-16-2 & VGG-8 & MobileV2 & MobileV2 & ShuffleV1  \\
			& & 70.50 $\pm$ 0.22 & 73.43 $\pm$ 0.22 & 70.46 $\pm$ 0.29 & 64.60 $\pm$ 0.32 & 64.60 $\pm$ 0.32 & 70.50 $\pm$ 0.22   \\
			\midrule
			\multirow{2}{*}{---} &\multirow{2}{*}{Teacher} & ResNet-32x4 & ResNet-32x4 & ResNet50 & WRN-40-2 & VGG-13 & WRN-40-2  \\
			& & 79.42 & 79.42 & 79.10 & 76.31 & 74.64 & 76.31  \\
			\midrule
			\multirow{4}{*}{Logits} & KD \cite{hinton2015distilling} & 74.00 $\pm$ 0.16 & 74.90 $\pm$ 0.29 & 73.81 $\pm$ 0.24 & 69.07 $\pm$ 0.26 & 67.37 $\pm$ 0.22 & 74.83 $\pm$ 0.13   \\
			& DTD-KA \cite{wen2021preparing} & 73.99 $\pm$ 0.12 & 74.11 $\pm$ 0.21 & 73.91 $\pm$ 0.21 & 68.99 $\pm$ 0.41 & 67.41 $\pm$ 0.12 & 74.90 $\pm$ 0.14   \\
			& VBD \cite{hegde2020variational} & 74.21 $\pm$ 0.21 & 74.32 $\pm$ 0.22 & 74.02 $\pm$ 0.22 & 69.22 $\pm$ 0.21 & 67.77 $\pm$ 0.21 & 75.10 $\pm$ 0.11   \\
			& DKD \cite{zhao2022decoupled} & 77.42 $\pm$ 0.11 & 76.68 $\pm$ 0.22 & 75.98 $\pm$ 0.22 & 69.47 $\pm$ 0.21 & 69.71 $\pm$ 0.26 & {\bf 76.41 $\pm$ 0.13}   \\
			\midrule
			\multirow{8}{*}{Features} & FitNet \cite{romero2014fitnets} & 74.82 $\pm$ 0.13 & 74.70 $\pm$ 0.35 & 73.72 $\pm$ 0.18 & 68.71 $\pm$ 0.21 & 63.16 $\pm$ 0.23 & 74.11 $\pm$ 0.23   \\
			& AT \cite{zagoruyko2016paying} & 74.76 $\pm$ 0.19  & 75.38 $\pm$ 0.18 & 73.45 $\pm$ 0.17 & 68.64 $\pm$ 0.12 & 63.42 $\pm$ 0.21 & 73.73 $\pm$ 0.19  \\
			& SP \cite{tung2019similarity} & 73.80 $\pm$ 0.21 & 75.16 $\pm$ 0.32 & 73.86 $\pm$ 0.15 & 68.48 $\pm$ 0.22 & 65.42 $\pm$ 0.21 & 74.01 $\pm$ 0.11   \\
			& VID \cite{ahn2019variational} & 74.28 $\pm$ 0.12 & 74.85 $\pm$ 0.35 & 73.75 $\pm$ 0.21 & 68.91 $\pm$ 0.21 & 65.70 $\pm$ 0.28 & 74.41 $\pm$ 0.26   \\
			& MGD \cite{yue2020mgd} & 75.34 $\pm$ 0.21 & 75.65 $\pm$ 0.08 & 73.98 $\pm$ 0.27 & 69.22 $\pm$ 0.24 & 66.23 $\pm$ 0.22 & 74.89 $\pm$ 0.22   \\
			& SRRL \cite{yang2021softmax} & 75.38 $\pm$ 0.31 & 75.46 $\pm$ 0.13 & 74.21 $\pm$ 0.20 & 69.34 $\pm$ 0.20 & 68.48 $\pm$ 0.31 & 75.22 $\pm$ 0.19 \\
			& CRD \cite{tian2019contrastive} & 75.46 $\pm$ 0.23 & 75.70 $\pm$ 0.29 & 74.42 $\pm$ 0.21 & 69.87 $\pm$ 0.17 & 69.73 $\pm$ 0.21 & 76.05 $\pm$ 0.23   \\
			& SemCKD \cite{chen2021cross} & 75.41 $\pm$ 0.11 & 75.65 $\pm$ 0.23 & 74.68 $\pm$ 0.22 & 69.88 $\pm$ 0.30 & 68.78 $\pm$ 0.22 & 74.81 $\pm$ 0.21  \\
			\midrule
			Logits & \name (Ours) & \textbf{77.88 $\pm$ 0.22} & \textbf{77.61 $\pm$ 0.21} & \textbf{76.49 $\pm$ 0.24} & \textbf{70.89 $\pm$ 0.14} & \textbf{71.02 $\pm$ 0.21} & {76.22 $\pm$ 0.11} \\
			\bottomrule
		\end{tabular}
	}
	\vspace{3pt}
	\caption{Top-1 test accuracy (\%) of the various distillation approaches of different architectures on CIFAR-100. Each experiment is repeated three times, and we report the mean and standard deviation of the top-1 accuracy.  %We denote the best result by Bold
    The best results appear in \textbf{bold}. }
	\label{Tbl:CIFAR-100-2}
\end{table*}  

\subsection{Main Results}\label{exp:main results}
%\noindent {\bf Setup.} 
\cref{Tbl:CIFAR-100-1} to \cref{Tbl:ImageNet} show a comprehensive performance comparison of various approaches based on student-teacher network combinations of various architectures, such as ResNet and VGG. 

\noindent {\bf Results on CIFAR-100.} %In \cref{Tbl:CIFAR-100-1} and \cref{Tbl:CIFAR-100-2},
In \cref{Tbl:CIFAR-100-1,Tbl:CIFAR-100-2}, we compare our \name with several prevalent distillation methods on CIFAR-100 based on two kinds of teacher-student network combination, \ie, the teacher-student pairs share similar architectures (ResNet110/ResNet-32, VGG-13/VGG-8) or heterogeneously architectures (ResNet-32$\times$4/ShuffleV1, VGG-13/MobileV2). As can be seen from the two tables, \name consistently outperforms other logit-based and feature-based distillation methods, achieving state-of-the-art performance.

%We also found some interesting phenomena. 
For comparison on similar architectures, logit-based distillation, except for DKD~\cite{zhao2022decoupled} and our \name, show inferior results to feature-based approaches; it may be owing to the flexibility of feature learning. However, when we switch the student-teacher pairs from uniform to heterogeneous architectures, the feature-based approaches show less superiority in homogeneous pairs, even sometimes underperform the vanilla KD. We ascribe this phenomenon to the feature-map gap between the student and different architectural teachers, which is hardly avoidable in feature distillation. Even in uniform architectural pairs, feature-based distillation achieves better performance at the expense of computational efficiency due to the inevitable computation of feature-map transformations, which may curb the industrial application of distillation. Our method surpasses the best feature-based distillation method without introducing feature map transformations, alleviating tedious computation without sacrificing performance.

\begin{table*}[t]
\setlength{\tabcolsep}{4.5pt}
\begin{center}
\begin{tabular}{l|cc|cccccc|c}
\toprule
 & Teacher & Student & AT & KD & SP & VID & MGD & CRD & \name \\
\midrule
Top-1 & 73.31 & 69.75 & 70.70 & 70.66 & 70.62 & 69.86 & 71.15 & 71.17 & \textbf{71.88} \\
Top-5 & 91.42 & 89.07 & 90.00 & 89.88 & 89.80 & 89.23 & 90.19 & 90.13 & \textbf{90.90}\\
\bottomrule
\end{tabular}
\caption{
Top-1 and Top-5 accuracies (\%) of student network ResNet-18 on the ImageNet validation set. We use ResNet-34 released by PyTorch official as the teacher network, and follow the standard training practice of ImageNet on PyTorch guideline.
}
\label{Tbl:ImageNet}
\end{center}
\vspace{-5pt}
\end{table*}
\noindent {\bf Results on ImageNet.} We conduct experiment using ResNet-34 as the teacher and ResNet-18 as the student. Similar results on CIFAR-100 also occur in the ImageNet experiment. As shown in \cref{Tbl:ImageNet}, our \name achieves the best classification performance on both Top-1 and Top-5 error rates compared with other existing distillation methods, which illustrates the robustness of our method on large-scale dataset learning.

%% file: body/discussion.tex
To better understand class-aware distillation, we conduct further experiments from four perspectives. First, we apply \name on several feature-based distillation methods and show our effectiveness on feature-based distillation. Second, we perform the ablation study to show the indispensability of our proposed loss. Then the sensitivity analysis is operated to figure out what influences the distillation. The experiments on other visual evaluations beyond classification tasks are shown in the Appendix, such as dense prediction and knowledge transfer. In Appendix, we also conduct visualization and training efficiency analysis to show the superiority of our \name over the state-of-the-art methods in performance and efficiency.

\subsection{Applicability on Feature Distillation}\label{dis:feature distillation}

To show the applicability of \name on feature-based distillation approaches, we embed our module in logit-based distillation to several mainstream feature-based distillation approaches and experiment on CIFAR-100 compared with the original feature-based distillation. From \cref{tbl:feature distillation}, we observe that all the feature-based distillation methods are improved when adding the feature-aware modules, and the highest enhancement occurs in the SRRL$_+$ framework. Hence, our approach encourages feature-based distillation without inducing extra computation costs.
\begin{table}[t]
\begin{center}
\begin{tabular}{l|cccc}
\toprule
 Metric (\%) & FitNet & AT & CRD & SRRL \\
\midrule
Accuracy & 75.75 & 75.28 & 76.01 & 75.55 \\
Accuracy$^+$ & {\bf 75.88} & {\bf 75.87} & {\bf 76.21} & {\bf 76.56} \\
\bottomrule
\end{tabular}
\caption{Comparison of classification accuracy between several mainstream feature distillation methods and the corresponding methods equipped with our feature-aware and feature correlation loss. The second row, Accuracy$^+$, shows the results when applying our approach on each feature distillation method. We use WRN-40-2 as the teacher and WRN-16-2 as the student. 
}\label{tbl:feature distillation}
\end{center}
\end{table}

\subsection{Ablation Study}
We conduct thorough ablation studies on \name from various views. We choose ResNet-32$\times$4 as the teacher for the following test. 
%We use ResNet-8$\times$4 as the student in the studies of different distribution loss, and extra distillation test will performed on ShuffleV1 in the ablation studies of and ShuffleV1 as the student distilled from ResNet-32$\times$4 as the teacher by default.

\begin{table}[t]
	\centering
	\resizebox{1\linewidth}{!}{
		\begin{tabular}{lcccccc}  
			\toprule
			\multirow{2}{*}{Module}& \multicolumn{4}{c}{Distillation}& \multirow{2}{*}{ResNet-8$\times$4} & \multirow{2}{*}{ShuffleV1}\\  
			& $L_{kl}$ & $L_{\NMSE}$ & $L_{cla}$ & $L_{dc}$ & & \\
			\midrule
			Baseline &-&-&-&-&72.51& 70.50 \\
			KD &\checkmark&-&-&-&74.12&74.00 \\
			w/o cla&-&\checkmark&-&-&76.11&75.12\\
			w/o cor&-&\checkmark&\checkmark &-&76.91&76.43\\
			\name &-&\checkmark&\checkmark&\checkmark&{\bf 77.68}&{\bf 77.88}\\
			\bottomrule
	\end{tabular}}
	\caption{Ablation study on the distillation loss on CIFAR-100. Baseline denotes the primary cross-entropy loss on the student model without knowledge distillation. And in other cases, the knowledge from pre-trained ResNet-32$\times$4 is used for distillation. }
	\label{tbl:ablation}
\end{table}

\subsubsection{Class-aware Knowledge and Correlation Loss}
We conduct the ablation study on the proposed loss using CIFAR-100 dataset, and the results are shown in \cref{tbl:ablation,tbl:loss comparison}. In \cref{tbl:ablation}, `KD' indicates we use Hinton's KD framework for the comparison group, which employs KL divergence to calculate the distribution discrepancy; `w/o cla' and `w/o cor' mean we use the proposed \NMSE loss to distill, but without the class distillation and the class correlation loss, respectively. We observe that in both two networks, \NMSE shows its superiority to KL divergence; in addition, the performance becomes even better with the exploitation of class-aware information and class correlation simulation, confirming the importance of the proposed \NMSE loss, class-aware information and class correlation loss. Similar results are also presented in \cref{tbl:loss comparison}, in which several prevalent metrics are employed to test the effect of class knowledge; thus, we conclude that the class information is beneficial for logit distillation.

\begin{table}[ht!]
\centering
\setlength{\tabcolsep}{2.5mm}{
\begin{tabular}{lcccc}
    \toprule
    \multirow{2}{*}{Loss} & \multicolumn{2}{c}{w/o class} & \multicolumn{2}{c}{w/ class}  \\
    & un-sup & sup & un-sup &sup \\
    \midrule
    KL & {73.61} & {74.12} & {74.00} & {74.68}\\
    JS-D\cite{menendez1997jensen} & {73.68} & 74.00 & {73.91} & {74.63} \\
    MSE & {74.51} & 75.88 & {75.12} & {76.22}\\
    $L_1$  & {73.78} & 74.22 & {74.49} & {74.83}\\
    \midrule
    \textbf{\NMSE} & {\bf 75.91} & {\bf 76.43} & {\bf 77.11} & {\bf 77.68}\\
    \bottomrule
\end{tabular}}
\caption{Accuracy of knowledge distillation framework on CIFAR-100 with different metrics to weight the discrepancy between the teacher and student logit losses and whether to use the class-level information. We use the ResNet-8$\times$4 network as the student. We also report the performance without the groundtruth labels, denoted as \textit{un-sup}.}
\label{tbl:loss comparison}
\end{table}

\subsubsection{Different Metric and Label-free Testing}\label{dis-abl:metrics}
From \cref{tbl:loss comparison}, we also observe that the proposed \NMSE loss outperforms other metrics for the calculation of discrepancy between the student and teacher in knowledge distillation. In addition, we also test the label-free learning in distillation which we cut down the cross-entropy loss. Even without the groundtruth labels, our performance only drops $0.53\%$ from $77.68\%$ to $77.11\%$. In contrast, the MSE results drop $0.8\%$ in the class-aware distillation and even worse in the `w/o class' condition, in which the MSE results drop $1.37\%$. Thus we can conclude that \name is less sensitive to the vacancy of labels.

%\subsection{Visualization}
\begin{figure} [t]
	\centering
	\subfloat[ResNet-32$\times$4$\longrightarrow$ResNet-8$\times$4]{
		\includegraphics[scale=0.38]{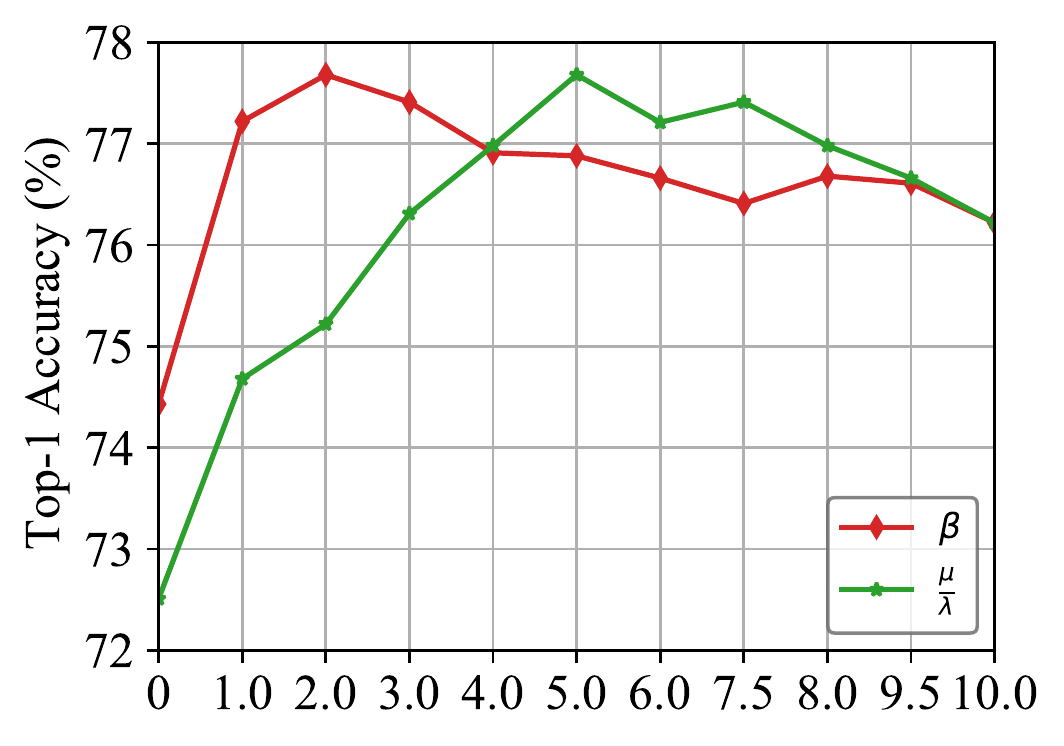}
		%\caption{$\lambda$ on CIFAR-100}
		}
	\subfloat[ResNet50$\longrightarrow$VGG-8]{
		\includegraphics[scale=0.38]{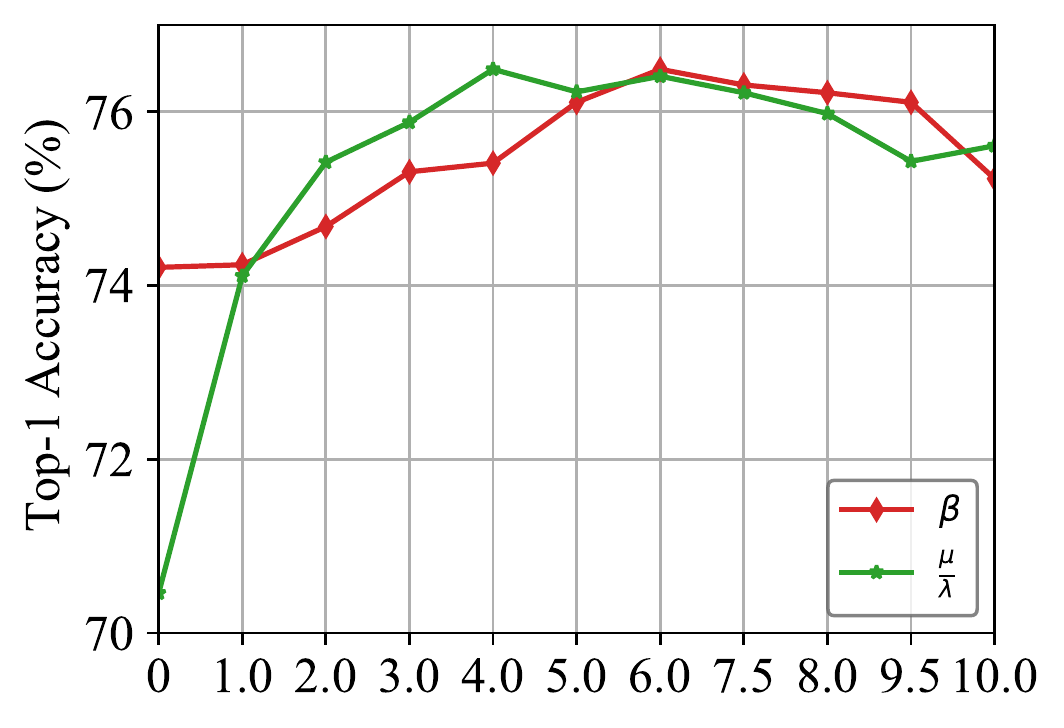}
		%\caption{$\lambda$ on ImageNet}
		}
	\caption{Sensitivity analysis for hyperparameters $\frac{\mu}{\lambda}$ and $\beta$.}
	\label{fig:sensitivity} 
\end{figure}

\subsection{Sensitivity Analysis}
%\subsubsection{Coefficient}
\noindent {\bf Analysis on Coefficient.} 
Here we conduct sensitivity analysis on the effect of hyper-parameter $\frac{\mu}{\lambda}$ and $\beta$ on the performance. We conduct experiments on CIFAR-100 dataset with $\frac{\mu}{\lambda}$ and $\beta \in \{0,1.0,2.0,3.0,4.0,5.0,6.0,7.0,8.0,9.0,10.0\}$. We set the teacher-student combination as ResNet-32$\times$4/ResNet-8$\times$4 and ResNet50/VGG-8 on CIFAR-100.  \cref{fig:sensitivity} shows the accuracy of each experiment with varying hyperparameters. We keep all settings the same as the basic setting in the main experiments except the testing hyperparameter. The results show that \name is robust to hyperparameter $\beta$ when $\beta \geq 1$. In addition, we notice that the worst performance happens when hyperparameter $\frac{\mu}{\lambda}$ or $\beta$ is set to $0$, demonstrating the indispensability of our class-aware distillation.

\begin{table}[t]
\centering
\small
\begin{tabular}{c|ccccc}
\toprule
Batch size & 64 & 128 & 256 & 512 & 1024  \\
\midrule
Acc. (\%) & 71.08 & 71.13 & 71.49 & {\bf 71.88} & 71.68  \\ 
\bottomrule
\end{tabular}
\vspace{.5em}
\caption{Effect of batch size on class distillation (ImageNet evaluation accuracy with 100-epoch pre-training with knowledge distillation from ResNet34 to ResNet18).
\label{tbl:batch}
}
\end{table}

%\subsubsection{Batch Size}
\noindent {\bf Analysis on Batch Size.} 
\cref{tbl:batch} reports the effect on the distillation performance with different batch size. One can observe that when the batch size rises from 64 to 512, the accuracy improves. This phenomenon lies on the ground that large batch size instances may provoke better class-level information. We also notice that the distillation does not work well when the batch is too large; with a 1024 batch, the performance is lower when we use a 512 batch. This phenomenon may due to the redundancy of class-instance similarity when the batch is too large. 

%\subsubsection{class Correlation}
\noindent {\bf Analysis on Class Correlation.} 
As shown in \cref{tbl:decor}, we may conclude that slight class correlation simulation encourages the student mimic the the teacher, which benefits the distillation. However, if the class correlation loss is too large, it would hinder the basic instance-level knowledge. 

\begin{table}[t]
\centering
\small
\begin{tabular}{c|cccccc}
\toprule
Coefficient & 0 & 0.2 & 0.5 & 1 & 5 & 10 \\
\midrule
Acc. (\%) & 76.68 & {\bf 77.68} & 77.31 & 77.11 & 76.61 & 76.72 \\%\underline{2.1} \\ 
\bottomrule
\end{tabular}
\vspace{.5em}
\caption{Effect of class correlation loss coefficient on distillation. (Distilling ResNet-8$\times$4 from ResNet-32$\times$4 on the CIFAR-100 dataset). The best result appears in {\bf bold}.% and \underline{underline} for worst result
\label{tbl:decor}
}
\end{table}

%% file: body/A.tex
\begin{table*}[!h]
\setlength{\tabcolsep}{2pt}
\begin{center}
\begin{tabular}{lccccccc|cc}
    \toprule 
    Model & \small{Baseline
    }& \small{DDGSD\cite{xu2019data}} & \small{BYOT\cite{zhang2019your}} & \small{CS-KD\cite{yun2020regularizing}} & \small{SLA+SD\cite{lee2020self}} & \small{FRSKD\cite{ji2021refine}} & \small{BAKE\cite{ge2021self}} & \small{\name} & $\Delta$\\
    \midrule 
    ResNet-50 & 76.80 & 77.10 & 77.40 & 77.61 & 77.20 & 76.68 & 78.00 & {\bf78.81} & +2.01\\
    ResNet-101 & 78.60 & 78.81 & 78.66 & 78.99 & 78.91 & 79.22 & 79.31 & {\bf79.91} & +1.31 \\
    \midrule 
    ResNeSt-50 & 78.40 & 78.66 & 78.60 & 78.71 & 78.98 & 78.91 & 79.31 & {\bf80.40} & +2.00 \\
    \midrule 
    ResNeXt-101 (32$\times 4d$) & 78.71 & 78.99 & 78.00 & 78.24 & 78.68 & 79.11 & 79.21 & {\bf80.12} & +1.41 \\
    \bottomrule
\end{tabular}
\caption{Comparison of self-distillation methods on ImageNet dataset using models of ResNet, ResNeSt and ResNeXt. We report the top-1 accuracy (\%) and the value in the last column are the performance improvement compared to vanilla classification. The best results on each architecture appear in {\bf bold}.}
\label{tbl:sd imagenet}
\end{center}
\vspace{-5pt}
\end{table*}

\begin{table*}[!h]
\centering
\begin{tabular}{l|c|cccccc|c}
\toprule
 & Student & KD & AT & FitNet & SRRL & \name & SRRL+ & Teacher\\
\midrule
CIFAR100$\rightarrow$STL-10        & 71.33 & 73.01 & 73.67 & 73.12 & 75.12 & \underline{76.24} & {\bf 76.67} & 70.60 \\
CIFAR100$\rightarrow$TinyImageNet & 35.10 & 35.39 & 35.42 & 35.55 & 37.13 & \underline{38.41} & {\bf 38.57} & 34.20 \\
\bottomrule
\end{tabular}
\caption{
\small{We conduct the experiment of feature transfer by using the representation learned from CIFAR100 to STL-10 and TinyImageNet datasets. We freeze the network and train a linear classifier on top of the last feature layer to perform 10-way (STL-10) or 200-way (TinyImageNet) classification. SRRL$+$ denotes that we extend our cluster-aware method on feature-based distillation SRRL. We use the combination of teacher network ResNet-32$\times$4 and student network ResNet-8$\times$4. The best result appears in {\bf bold}, and the second best is \underline{underlined}.
}
}
\label{tbl:transfer}
\end{table*}
\section{Self-Distillation Results.} 
To further demonstrate that \name may also improve the teacher-free distillation, we use the teacher-free framework proposed in CS-KD~\cite{yun2020regularizing} and replace the original loss in CS-KD with our loss as in Eq.~\ref{eq:loss}. We evaluate CLKD on ImageNet with several prevailing teacher-free distillation methods. As shown in \cref{tbl:sd imagenet}, our approach exhibits higher performance than other self-knowledge distillation baselines on the ImageNet dataset. We not only consider the prevailing ResNet architectures (\eg, ResNet-50), but also test on the ResNeSts~\cite{zhang2022resnest} and ResNeXts~\cite{xie2017aggregated} networks. And the accuracy on ImageNet is improved by 2.01\%, 1.31\% on ResNet-based architectures. We also achieve similar enhancement with ResNeSts and ResNeXts, which show that our approach still work in the teacher-free paradigm.

\section{Feature Transferability}
We continue to conduct several experiments to examine the feature transferability of \name.
As shown in \cref{tbl:transfer}, we train linear fully-connected (FC) layers as the classifier with the feature extractor frozen for STL-10 and Tiny-ImageNet datasets. We use an SGD optimizer with 0.9 momentum and no weight decay strategy in classifier training. We set the batch size to 128, and the number of total epochs is 40. Our initial learning rate is set to 0.1, then divided by 10 for every 10 epochs. From \cref{tbl:transfer}, we observe that our method beats all existing techniques, manifesting its feature transferability. Besides, the best results occur when we use \name on SRRL, which further demonstrates its extensibility on feature distillation.

\begin{table*}[!h]
\begin{center}
\begin{tabular}{l|cc|cccccc}
\toprule
 Metric (\%) & Teacher & Student & KD & AT & FitNet & CRD & SKD\cite{liu2019structured} & CLKD \\
\midrule
mIoU & 78.56 & 69.10 & 68.70 & 69.71 & 69.99 & 71.51 & {\bf 72.54} & \underline{72.22} \\
\bottomrule
\end{tabular}
\caption{The segmentation performance comparison on Cityscapes~\cite{cordts2015cityscapes} val dataset. Teacher: ResNet101 and Student: ResNet18. The mIoU measures the mean intersection of the prediction and the groundtruth pixels.The best result appears in {\bf bold}, and the second best is \underline{underlined}.
}\label{tbl:structural distillation}
\end{center}
\end{table*}

\begin{table}[!h]
\begin{center}
\begin{tabular}{l|cccc}
\toprule
 Method  & Accuracy (\%) & Time (s) & Extra params\\
\midrule
KD & 73.33 & {\bf 12} & {\bf 0} \\
AT & 74.57 & 13 & {\bf 0} \\
FitNet & 73.89 & 14 & 16.8K \\
CRD & 75.59 & 35 & 12.3M \\
SemCKD & 75.58 & 33 & 12.1M \\
\midrule
CLKD & {\bf 77.68} & {13} & {\bf 0} \\
\bottomrule
\end{tabular}
\caption{Training time~(per batch), extra parameters versus accuracy on CIFAR-100.
Teacher: ResNet-32$\times$4, student: ResNet-8$\times$4.
}\label{tbl:training efficiency}
\end{center}
\end{table}
% \begin{table}[h]
%     \centering
%     \begin{tabular}{l|ccccc|c}
%     \toprule
%         Method & KD & AT & FitNet & CRD & SemCKD & CLKD  \\
%          & 
%     \end{tabular}
%     \caption{Caption}
%     \label{tab:my_label}
% \end{table}

\section{Structured Knowledge Distillation}
Unlike classification, semantic segmentation aims at dense pixel prediction, thus requires the holistic understanding of structured context across pixels. To show that our approach \name may learns semantic structural knowledge, we conduct several knowledge distillation test on dense pixel tasks---semantic segmentation. As shown in \cref{tbl:structural distillation}, conventional logit-based knowledge distillation dealing with structural prediction tasks may not achieve a desirable result and even has a deteriorating impact. This is because only considering instance-level information may lead to a loss of structured context among pixels. We observe the best performance occurs when using SKD~\cite{liu2019structured}, which is designed for segmentation. And our approach underperforms SKD only slightly and outperforms all the KD approaches designed for classification. Therefore our method captures the structural information in the logit distillation framework. 

\begin{figure} [t]
	\centering
	\subfloat[SRRL\cite{yang2021softmax}]{
		\includegraphics[scale=0.28]{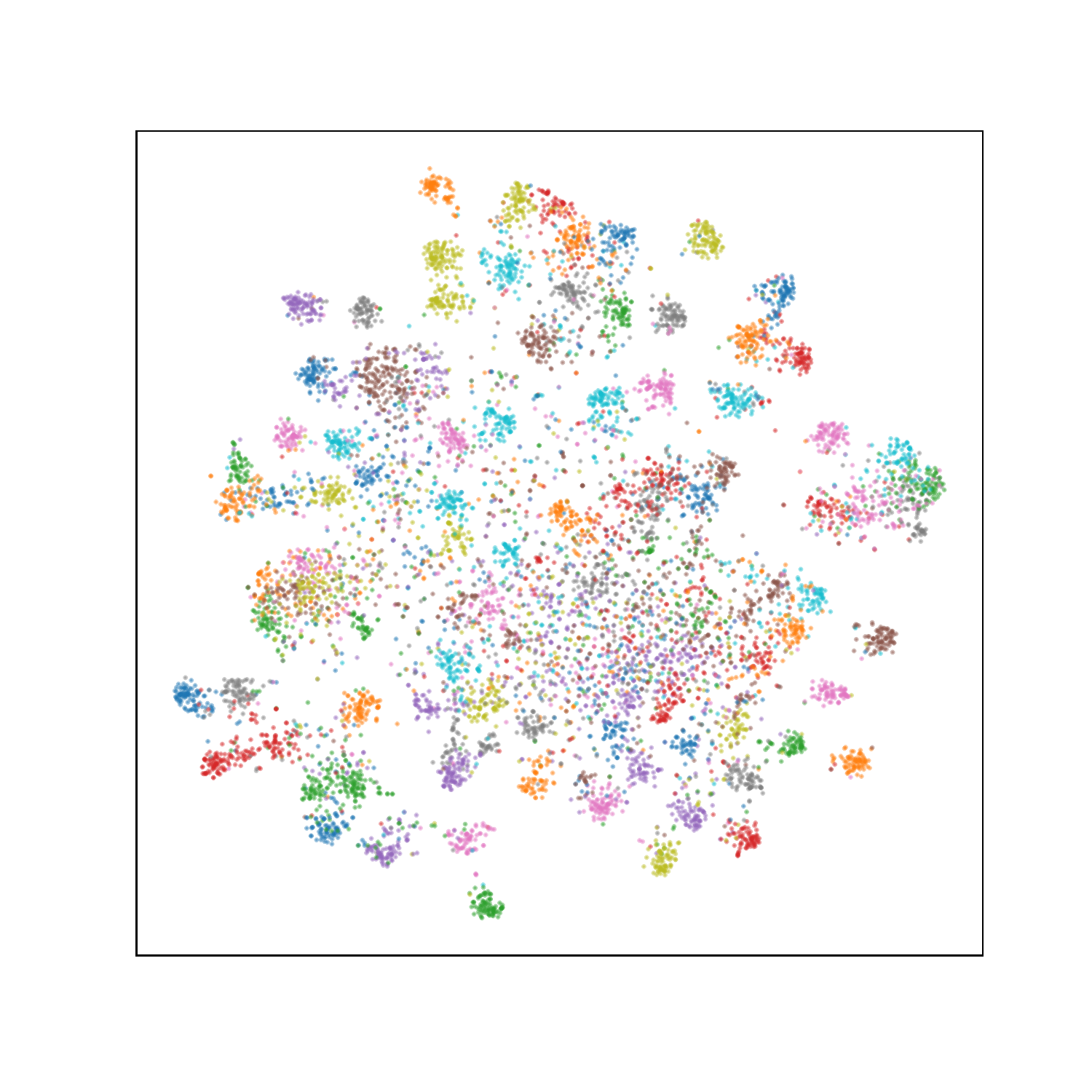}
		%\caption{$\lambda$ on CIFAR-100}
		}
	\subfloat[CRD\cite{tian2019contrastive}]{
		\includegraphics[scale=0.28]{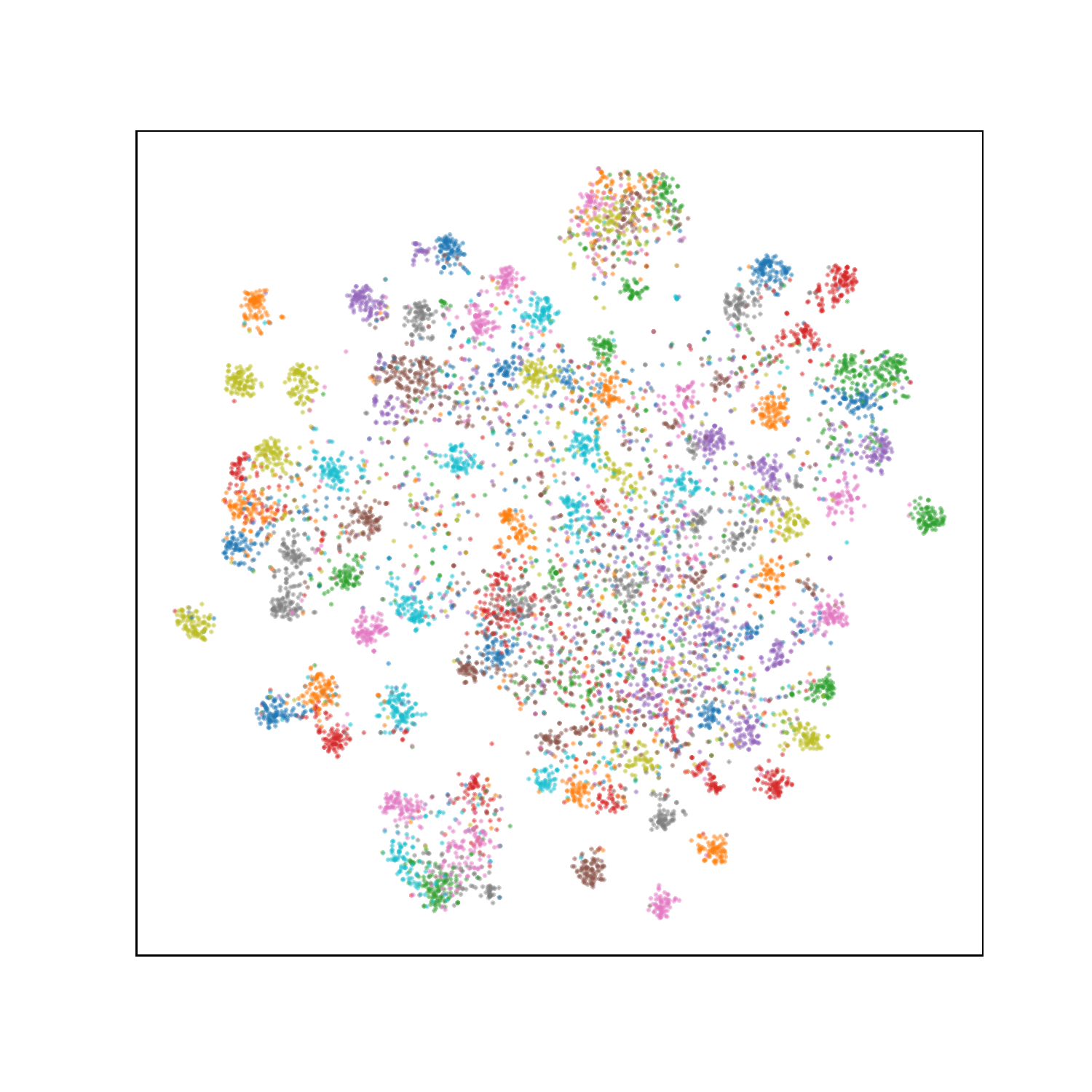}
		%\caption{$\lambda$ on ImageNet}
		}
	\\
	\subfloat[KD\cite{hinton2015distilling}]{
		\includegraphics[scale=0.28]{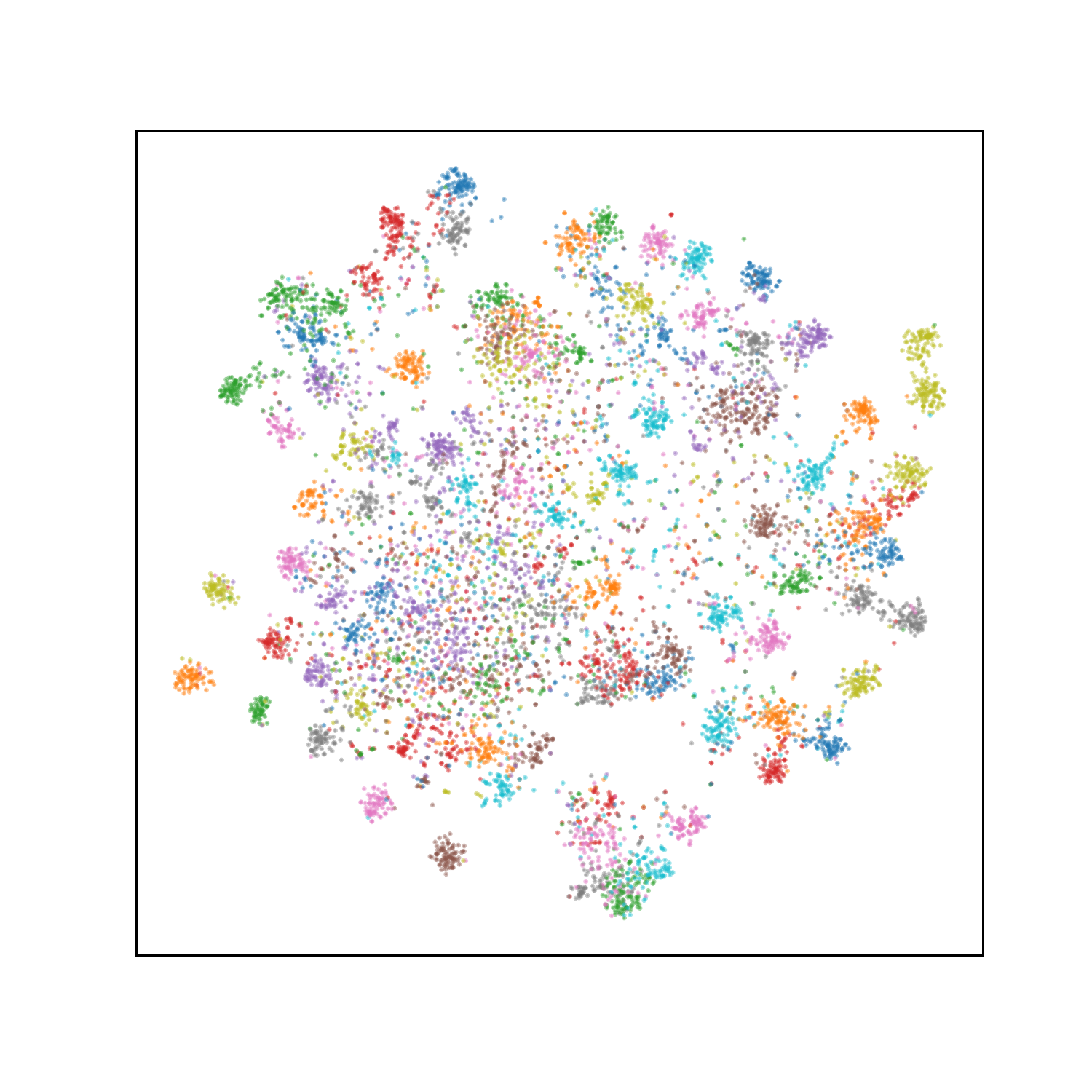}
		%\caption{$\beta$ on CIFAR-100}
		}
	\subfloat[Our \name]{
		\includegraphics[scale=0.28]{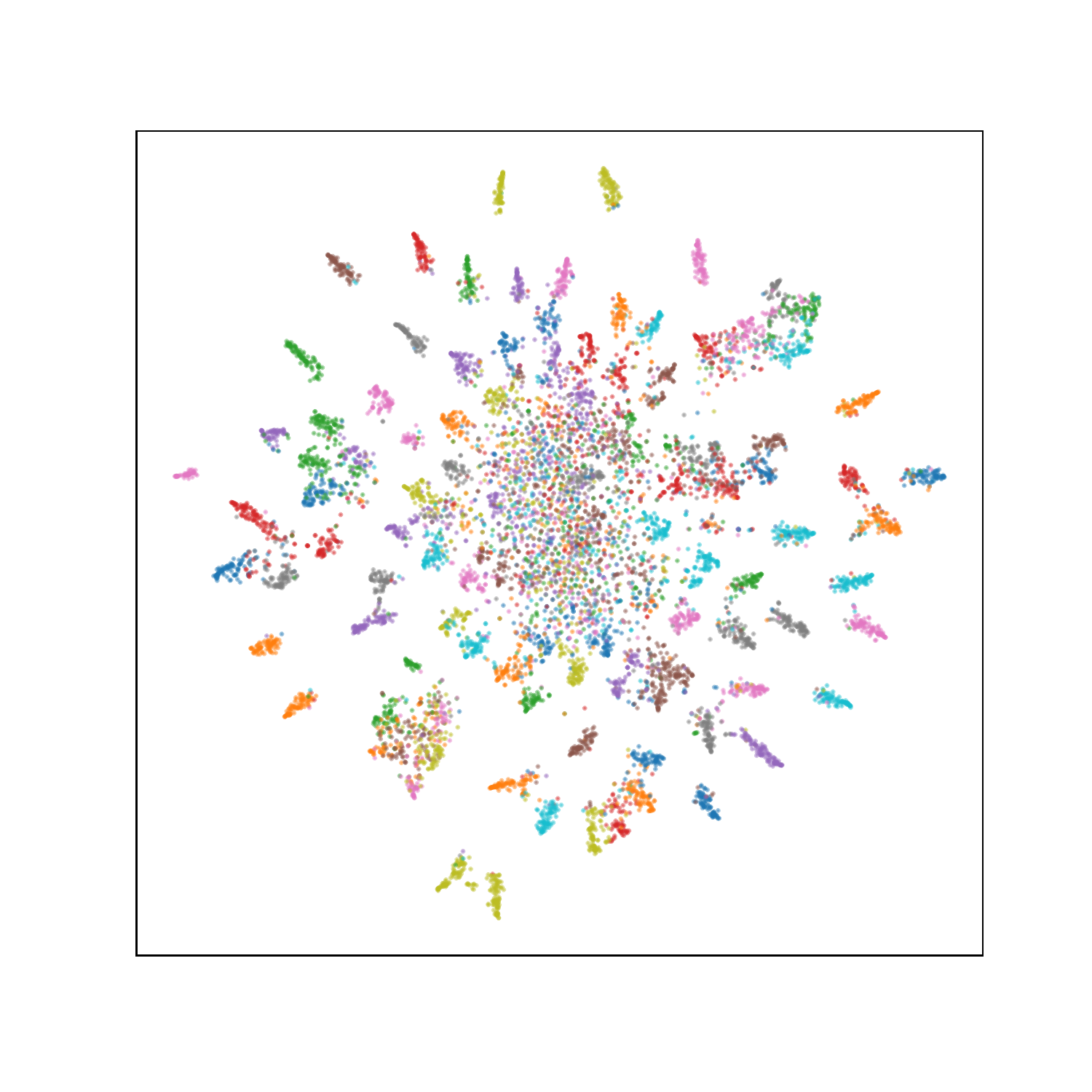}
		%\caption{$\beta$ on ImageNet}
		}
	\caption{T-SNE visualization of feature learned from ResNet-8$\times$4 on the CIFAR-100 dataset with the knowledge distilled from ResNet-32$\times$4. We use feature before the classifier module for visualization.}
	\label{fig:tsne} 
\end{figure}
\section{Training Efficiency}
%To manifest the training efficiency of \name, we compare its training time with several existing distillation methods. 
We evaluate the training efficiency of our approach with several existing distillation methods. As shown in \cref{tbl:training efficiency}, \name achieves the best performance without extra training consumption~(\eg, training time and extra parameters). Therefore our \name outperforms all existing feature-based distillation methods in performance and efficiency, and may be applied in practicality.
\section{Visualization}
We present t-SNE visualizations of several existing distillation methods and our \name. From Fig.~\ref{fig:tsne}, we may observe that our \name representations show better separability compared with other distillation methods such as CRD, SRRL, and vanilla KD. This result confirms that our class-aware distillation learns more discernable features and benefits the classification performance.